# GenNav: A Generic Indoor Navigation System for any Mobile Robot

Sudarshan S Harithas
Department Electronics and Communication Engineering
BMS College of Engineering
Bangalore , India
1bm16ec109@bmsce.ac.in

Biswajit Pardia
Zen Aerologiks
Bangalore ,India
biswajit@aerologiks.com

*Abstract*—The navigation system is at the heart of any mobile robot it comprises of SLAM and path planning units, which is utilized by the robot to generate a map of the environment, localize itself within it and determine an optimal a path to the destination. This paper describes the conceptualization, development, simulation and hardware implementation of GenNav a generic indoor navigation system for any mobile aerial or ground robot. The generalization is brought about by modularizing and creating independence between the software computation and hardware actuation units by providing an alternate source of source of odometry from the LiDAR eliminating the requirement for dedicated odometry sensors. The odometry feedback from the LiDAR can be used by the navigation computation unit and the system can be generalized to a wide variety of robots, with different type and orientation of actuators

## I. INTRODUCTION

Autonomous mobile robots are increasingly becoming prevalent in recent times owing to their applications in a multitude of sectors. The navigation system is of central importance to any mobile robot, it performs two interconnected tasks of SLAM and Path-Planning that work synchronously with the hardware.

The hardware actuators are continuously controlled and their feedback is monitored through the Navigation system software. This paper deals with the implementation of GenNav a generic navigation system which can be implemented on any mobile aerial or ground robot for indoor navigation. The working of GenNav is independent of the type (Dc, Servo, stepper, BLDC motor etc.), orientation and position of actuators used for the locomotion of the robot. The idea has been previously explored in the form of GeRoNa [2] which is generic navigation framework for wheeled robots, GenNav proposes a novel approach for navigation which is not limited to wheeled robots but can also be extended aerial vehicles.

The performance of GenNav is determined by the independence, modularization and abstraction between the hardware actuators and software computation unit. There can be two approaches to achieve this: Firstly, an open-loop approach can be followed where the feedback from the actuators are disconnected or a closed loop approach where the odometry feedback is obtained through an alternate source from the existing on-board scanning device such as LiDAR or RGB-D camera. The closed loop approach is the preferred choice as the former is susceptible to noise and is unreliable.

ROS (Robot Operating System) has been extensively used in the development of GenNav. Simulations have been conducted on ROS supported simulators such as Rviz and Gazebo, the simulations have preceded hardware development and GenNav has been implemented on a custom designed differential drive robot and on a quadrotor. The robots that are used for during simulation is shown in figure 1.

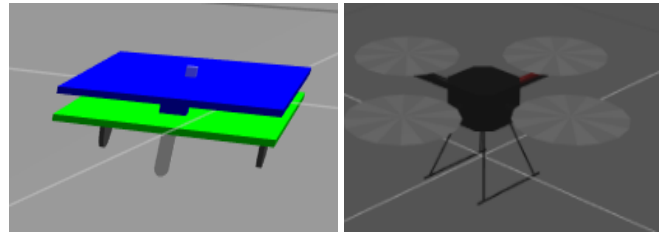

*Fig1: Mobile Robots (Differential Drive robot and quadrotor) on which GenNav has been implemented.*

## II. ARCHITECTURE OF A MOBILE ROBOT

This section deals with the conceptualization and development of the GenNav. A detail explanation of the integration, data flow and working of various layers in the navigation system is provided below

### A. MECHATRONIC ARCHITECTURE

The integration and data flow from the sensors through the computation and embedded system units to the final actuation of the motors is described by its mechatronic architecture as depicted in figure 2.

The hardware abstraction layer includes sensors and its corresponding device drivers. The sensors used must be capable of providing a registered depth map and create a 3D view of the environment in which it operates. Sensors such as LiDAR or a RGB-D camera are utilized in this layer.

An Ubuntu Operating System with ROS form the platform for the computation layer. It is concerned with the task of sensor data collection, processing, navigation (SLAM and path planning) [7] program execution and actuation control command delivery to the embedded system

The actuation control for locomotion is performed by a microcontroller through the motor driver. The actuation control layer processes the commands from the computation layer and actuates the motors for locomotion. In a traditional setup of the navigation system this layer also acts as a feedback channel passing odometry data from the actuators to the computation layer.

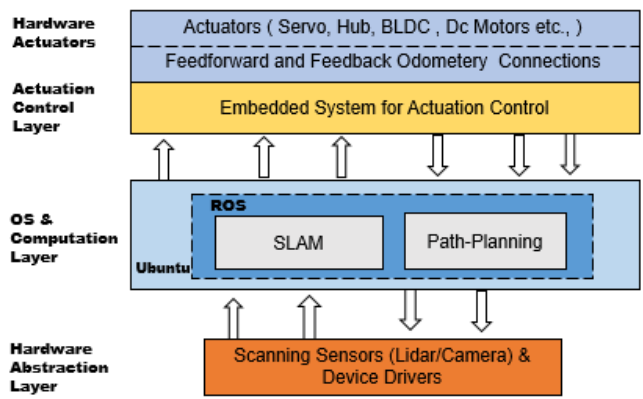

Fig2: Mechatronic Architecture of a Navigation system in a Mobile Robot

With GenNav, the actuation control layer is used only to perform locomotion control and is not concerned with facilitating feedback as the odometry is obtained from the scanning device and the computational layer remains unchanged.

*B. SOFTWARE ARCHITECTURE*

The software layer comprises of the programs to perform SLAM and path-planning as shown in figure 3, it is a part of the "OS and Computation Layer" in the mechatronic architecture

The structural arrangement of the robot during simulation is given in the form of a URDF (Unified Robot Description Format) file. The Tf describes the geometrical transform between the frames of reference of various on board sensor and actuators. The LiDAR, robot base and odometry are the frames of reference that are present of the robot. The Tf (transforms) and the LiDAR scan data are inputs to the software computation unit.

SLAM (Simultaneous Localization and Mapping) is the process of generating a Map of the environment in which the robot operates and simultaneously localizing itself within the generated map. The process of mapping requires human supervision, a human operator would guide the robot through the environment and the LiDAR scan data would be used to generate the map of the surroundings. Localization is the task of determining the position of the robot relative to the existing map [7].

Gmapping is used to perform mapping it deploys a particle filter based approach along with Rao-Blackwellized Particle Filters [3]. Localization within the map is implemented using AMCL (Adaptive Monte Carlo Localization) [6] it is a particle filter localization technique.

The Laser Scan Data is converted to odometry using the range flow approach [1] where the motion estimate is obtained by minimizing the robust function and the range flow constraint equation is determined in terms of sensor velocity. Through this approach an alternate source of odometry is obtained from the LiDAR and the computation layer is not dependent for the feedback from hardware sensors such as motor encoders and IMU.

The goal of Path-Planning is to determine an optimal path from source to destination. ROS provides for two path planners, a Global and a Local Path Planner. The global path planner [10] is concerned with the task of generating an optimal path from the existing position to the destination. Avoiding the obstacles within the immediate surrounding of robot is performed by the local path planner [9], this ensures that the robot can safely navigate between obstacles without collision

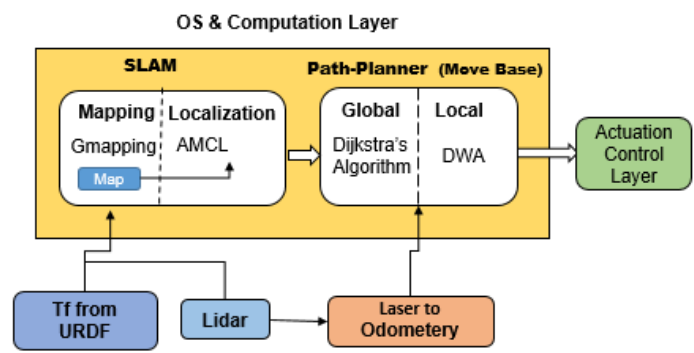

Fig3 : Software Architectue

The results of the SLAM process which include the map of the environment and the localization result along with the odometry data is used by the local and global planners to perform path-planning. Dijkstra's algorithm is used as the global planner [10] and DWA (Dynamic Window Approach) for the local path planner [9]. The planners are supported with "costmaps", which are maps where the boundaries of the obstacles are inflated. The global and local costmaps are used by the global and the local planners respectively. The costmaps help in generating a collision free path-plan and ensures optimal route from source to destination.

The integration of the global and local planners with its corresponding costmaps is performed by move base within ROS Navigation stack. The result of this computation process is the actuation signal which is a velocity command that would be fed into the Actuation control Layer

*C. ACTUATION CONTROL ARCHITECTURE*

The intention of the actuation control system is to ensure the delivery the target actuation velocity output equal to the reference velocity command.

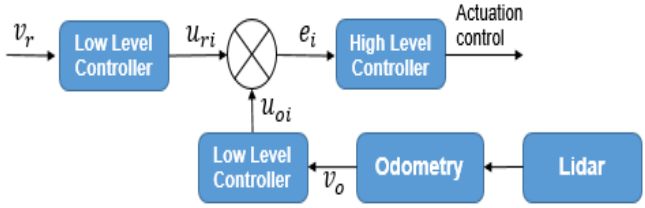

*Fig4: Actuation control Architecture*

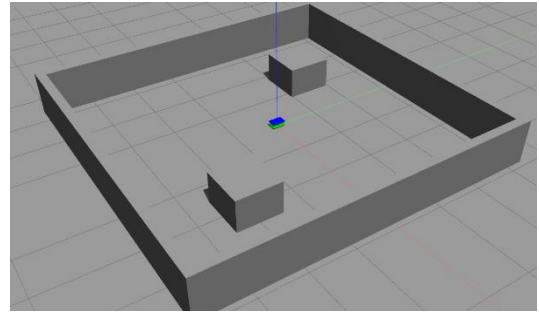

*Fig5. Differential drive robot in the gazebo World.*

Fig 4, describes the working of the actuation control layer. Here we have implemented a twostep actuation control sequence comprising of the High Level and Low Level controllers. We can observe that the feedback route consists of the Lidar sensor and the odometry is obtained through Laser scan to odometry conversion.

The locomotion of the robot in space can be governed by its kinematic and dynamic models, these models are implemented in the Low Level controller. The value $v_r$ is the velocity command obtained from the computation layer and $v_o$ is the actual velocity of the robot obtained from the odometry data.

The low level controller implements inverse kinematics/dynamics to convert the velocities in robot reference frame to equivalent induvial actuator input $u_{xi}$ where $i$ refers to the $i^{th}$ actuator i.e. $i$ takes 2 values in case of differential drive robot and 4 values in case of a quadrotor. Within GenNav the Low Level Controller is the only unit which is specific to the hardware actuator of the robot.

The error $e_i$ is computed for induvial actuator and this is passed into the PID controller which in-turn passes the actuation control signal into the $i^{th}$ actuator.

## III. IMPLEMNTATION

This section demonstrates the working of GenNav on a differential drive robot and a quadrotor these robots are chosen as they have different orientation and actuation control mechanism.

### A. Navigation of differential drive robot using GenNav : Simulation Results

ROS provides for simulators such as Rviz and Gazebo, the process of simulation was started with the generation of URDF and importing it onto ROS. Gazebo is a simulator which represents the real world robot model and is used to implement and test various algorithms it is the environement in which the robot operates on course of the simulation. Rviz is a visualizer which provides graphical tools for visualization and interaction with the robot

As seen in figure 5 a gazebo world is constructed with two blocks in and a surrounding wall. The process of implementation of GenNav for the navigation of a differential drive robot is shown in figure 6.

The SLAM process is initiated where the robot moves around the environments with human assistance generating the map of the environment and localized itself within the given map. The map during the course of generation can be visualized in Rviz as shown in figure 6 (a). Throughout the SLAM process the odometry data is obtained from the onboard LiDAR, using the using the Range Flow approach [1].

Fig6 (B) depicts the implementation of AMCL for localization of the differential drive robot. The pose array estimate can be observed beneath the robot which represents the possible position and orientation of the robot. The accuracy of the localization estimate can be by moving the robot within the environment. The costmaps which are inflated obstacle maps can also be observed in the image.

Figure 6 (b) depicts the implementation of AMCL for localization of the differential drive robot. The pose array estimate can be observed beneath the robot which represents the possible position and orientation of the robot. The accuracy of the localization estimate can be increased by moving the robot within the environment. The costmaps which are inflated obstacle maps can also be observed in the figure.

Path planning is implemented using move base within the ROS navigation stack .The path to the defined goal is planned using Dijkstra's algorithm [10] the global path planner and DWA (Dynamic Window Approach) as the local planner. The brown line in figure 6 (c) represents the generated path plan from source to destination.

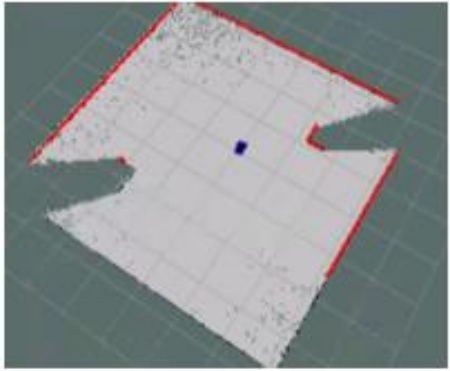
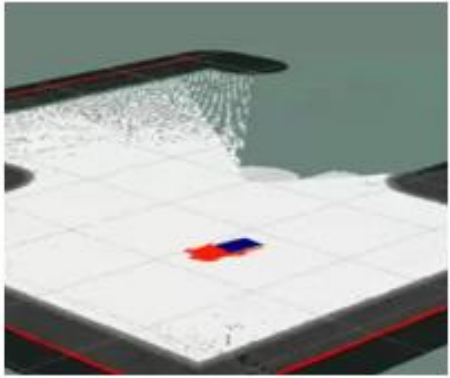
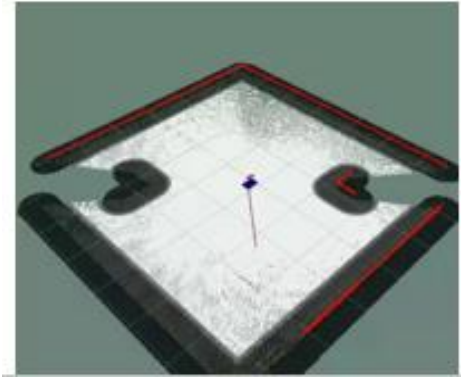

*Fig6: Implementation of GenNav on a Differential Drive Robot*

### B. Navigation of UAV using GenNav: Simulation Results

The quadrotor shown in figure 1, is used to demonstrate the implementation of GenNav on a UAV during simulation. The presence of the quadrotor in the Gazebo world can be seen in figure 7.

GenNav can be implemented on a UAV with a similar process to a differential drive robot with changes in few of the parameters given in [8] and the low level controller. The drone is moved around the environment to generating a map using Gmapping as shown in figure 8 (a). The Generated Map is used for Localization using AMCL as depicted in figure 8 (b). Path planning is implemented using move base the results are shown in figure 8 (c).

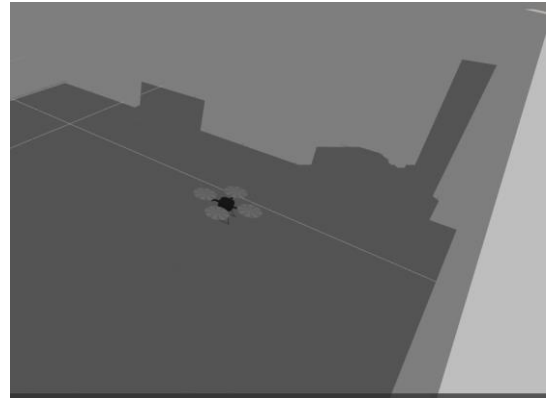

*Fig7: quadrotor in the Gazebo World*

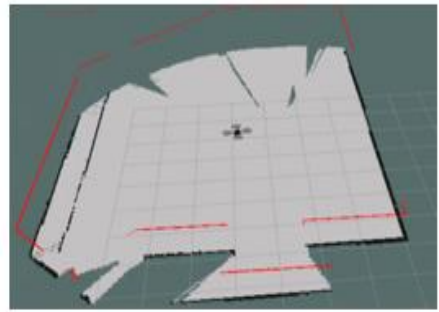
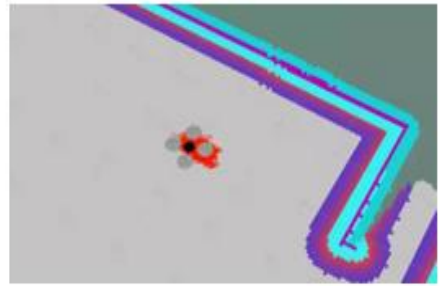
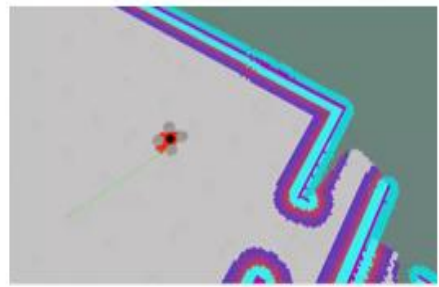

*Fig8: Implementation of GenNav on a quadrotor*

### C. GenNav on differential drive robot: Hardware Implementation

After the successful simulation we proceeded towards the hardware implementation, where GenNav is implemented on a differential drive robot shown in figure 9 with DC motors without encoders. An RPLIDAR A2M8 is mounted on the robot and a Jetson TX2 is used for computation and the ATmega2560 (Arduino Mega) is the microcontroller used.

The robot is powered with a 14.8V 22Ah LiPo battery through the power distribution circuit.

The steps followed during simulation were repeated for the hardware implementation, the robot mapped the indoor space in which it should operate using Gmapping, AMCL localization was implemented and move base was used for path planning. Figure 10 shows the resulting map which was generated and the goal position being published through Rviz using an interactive maker.

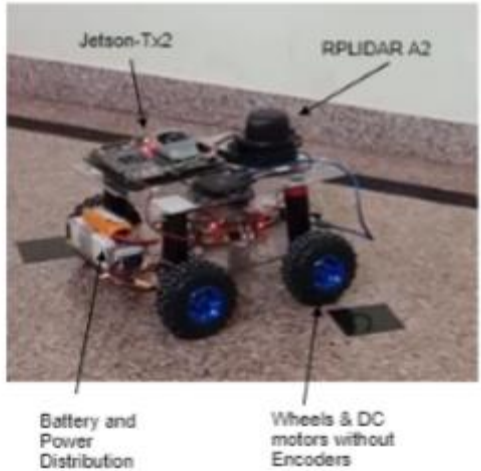

*Fig 9: Robot Constructed for demonstration*

The steps followed during simulation were repeated for the hardware implementation, the robot mapped the indoor space in which it should operate using Gmapping, AMCL localization was implemented and Move Base was used for path planning. Fig 10 shows the resulting map which was generated and the goal position being published through Rviz using an interactive maker.

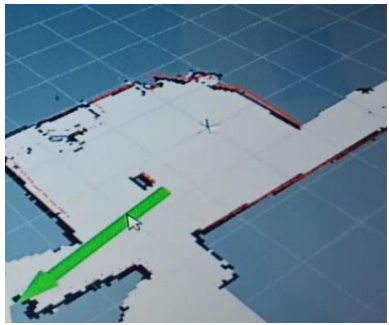

Fig10: Setting the destination position in the generated Map.

Once the destination is set the path is computed and traversed by the robot by providing the necessary velocity commands through the High Level PID Controller to the DC motors. Figure 11, shows the final point which the robot has reached once it completed the path traversal.

**Battery Monitoring and PWM Correction**

During hardware implementation it was observed that the performance of the robot was severely affected by the decrease in the battery voltage, this was undesirable as it reduced accuracy and the usage period of the robot. A battery monitoring system was designed that would determine the present voltage levels of the battery and appropriately correct the PWM output of the PID controller.

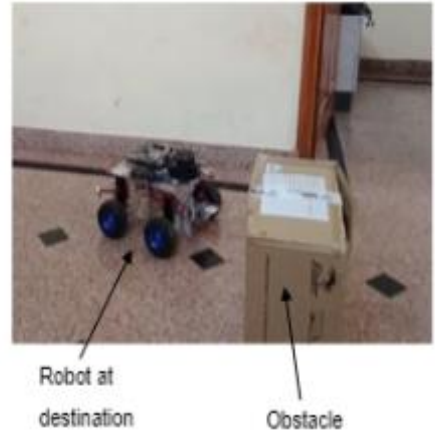

*Fig11: Robot reaching the goal*

Let us consider a robot which is expected to move with a velocity u at a PWM value of k, given the voltage level of the battery is V then equation 1 provides the necessary equation relating the above quantities. Where γ is a constant of proportionality.

$$u = \gamma.k.V \quad (1)$$

Let the present battery voltage reduce form $V$ to $V'$ whose difference is given by $\Delta V$, to compensate for this change the PWM must increase form $k$ to $k'$ whose difference is given by $\Delta k$. These changes are represented in equation (2).

$$\Delta u = \gamma\,[\,\Delta k.V + \Delta V.k\,] \quad (2)$$

The PWM must be altered such that the velocity does not change, given a change in voltage of the battery V, hence $\Delta u = 0$. Writing equation 2 with the above substitution we get.

$$\begin{aligned}0 &= \gamma\,[\,\Delta k.V + \Delta V.k\,]\\ [\,\Delta k.V &+ \Delta V.k\,] = 0\end{aligned} \quad (3)$$

Substituting values of $\Delta k$ and $\Delta V$ we get:

$$(k' - k).V = -(V' - V).k$$

$$(k' - k) = -(V' - V).\frac{k}{V}$$

$$k' = k - (V' - V) \cdot \frac{k}{V}$$

$$k' = k \left(1 - \frac{(V' - V)}{V}\right)$$

$$k' = k \left(\frac{2V - V'}{V}\right) \qquad (4)$$

Equation 4 provides the modified PWM value such that the robot moves with a constant velocity, given a change in battery voltage. This correction incorporated into the actuation control layer.

Using GenNav with the battery monitoring and PWM correction the robot could navigate to any point with a maximum error of 10cm, this error is the Euclidean distance (de) between the center of the robot and the destination point, an angular error (α) of less than 200 was obtained. Figure 12, depicts the measurement convention used while estimating the error. The ideal pose denotes the expected position and orientation and the practical result was the actual position and orientation result that was obtained. All measurements were performed taking the center of the robot as the point of reference.

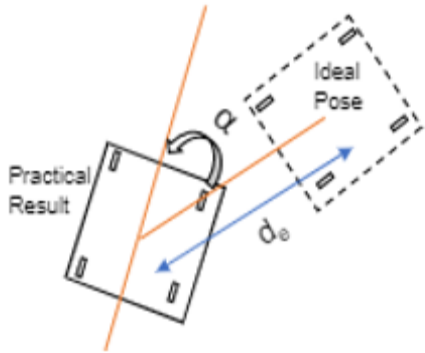

*Fig 12: Error Measurement Convention used*

## IV. CONCLUSION

The conceptualization, development of a generic indoor navigation system called GenNav has been detailed in this paper. The working of GenNav is tested through simulation and hardware implementation. During simulation GenNav has been successfully implemented on a differential drive robot and on a quadrotor. A custom designed differential drive robot is used to measure the accuracy of the navigation system in hardware it was found to have a maximum spatial and angular error of 10cm (Euclidean distance) and 200 respectively.